\documentclass{article}

\usepackage{arxiv}

\usepackage[utf8]{inputenc} % allow utf-8 input
\usepackage[T1]{fontenc}    % use 8-bit T1 fonts
\usepackage{hyperref}       % hyperlinks
\usepackage{url}            % simple URL typesetting
\usepackage{booktabs}       % professional-quality tables
\usepackage{amsfonts}       % blackboard math symbols
\usepackage{nicefrac}       % compact symbols for 1/2, etc.
\usepackage{microtype}      % microtypography
\usepackage{graphicx}
\usepackage{natbib}
\usepackage{doi}
\usepackage{amsmath}
\usepackage{xcolor}
\usepackage[most]{tcolorbox}
\usepackage{geometry}
\usepackage{enumitem}
\usepackage{lmodern}
\usepackage{fancyhdr}
\usepackage{subcaption}

\hypersetup{
    colorlinks=true,
    linkcolor=teal,
    filecolor=magenta,      
    urlcolor=cyan,
    citecolor=blue,
    pdftitle={Gazal-R1: Achieving State-of-the-Art Medical Reasoning with Parameter-Efficient Two-Stage Training},
    pdfpagemode=FullScreen,
}

\definecolor{boxbgcolor}{HTML}{F0F8FF} % AliceBlue
\definecolor{boxframecolor}{HTML}{4682B4} % SteelBlue
\definecolor{badcolor}{HTML}{FFE4E1}    % MistyRose
\definecolor{badframe}{HTML}{B22222}    % FireBrick
\definecolor{goodcolor}{HTML}{E0FFE0}   % Honeydew
\definecolor{goodframe}{HTML}{228B22}   % ForestGreen

\newtcolorbox{infobox}[1][]{
    colback=boxbgcolor,
    colframe=boxframecolor,
    fonttitle=\bfseries,
    coltitle=black,
    sharp corners,
    boxrule=1pt,
    #1
}

\newtcolorbox{promptbox}[1][]{
    colback=gray!5,
    colframe=gray!75,
    fonttitle=\bfseries\sffamily,
    coltitle=black,
    sharp corners,
    boxrule=0.5pt,
    fontupper=\small\ttfamily,
}

\newtcolorbox{observationbox}[1][]{
    colback=obsbgcolor,
    colframe=obsframecolor,
    fonttitle=\bfseries,
    coltitle=black,
    sharp corners,
    boxrule=1pt,
    #1
}

\title{Gazal-R1: Achieving State-of-the-Art Medical Reasoning with Parameter-Efficient Two-Stage Training}

\author{\hspace{1mm}Ahmed M.~Adly\\
	Research Engineer\\
	TachyHealth\\
        Riyadh 13316, Saudi Arabia \\
	\texttt{amostafa@tachyhealth.com} \\
	\And
	{\hspace{1mm}Mostafa~Samy} \\
	Data Science Product Manager\\
	TachyHealth\\
        Riyadh 13316, Saudi Arabia \\
	\texttt{msamy@tachyhealth.com} \\
        \And
        {\hspace{1mm}Amr~Fawzy~\thanks{Dr. Fawzy's profound expertise as a medical practitioner played a pivotal role in the comprehensive medical validation and stringent verification process of the generated responses, affirming their scientific integrity and practical utility.}} \\
	Chief Medical Officer\\
	TachyHealth\\
        Riyadh 13316, Saudi Arabia \\
	\texttt{amr@tachyhealth.com} \\
}

\renewcommand{\shorttitle}{\textit{Technical Report}}

\begin{document}
\maketitle

\begin{abstract}
\noindent We present Gazal-R1, a 32-billion-parameter language model that achieves state-of-the-art performance in medical reasoning while providing transparent, step-by-step explanations for clinical decision-making. Built upon Qwen3 32B, our model demonstrates that strategic training can enable mid-sized models to outperform significantly larger counterparts in specialized domains. We developed a novel two-stage training pipeline: first, supervised fine-tuning on a carefully curated dataset of 107,033 synthetic medical reasoning examples that teaches structured clinical thinking, enhanced by advanced parameter-efficient techniques including Weight-Decomposed Low-Rank Adaptation (DoRA) and Rank-Stabilized LoRA (rsLoRA); second, reinforcement learning using Group Relative Policy Optimization (GRPO) with a sophisticated multi-component reward system that refines accuracy, format adherence, and reasoning quality. Gazal-R1 achieves exceptional performance across medical benchmarks, scoring 87.1\% on MedQA, 81.6\% on MMLU Pro (Medical), and 79.6\% on PubMedQA, surpassing models up to 12× larger. Beyond its strong empirical results, this work provides detailed insights into the challenges of training reasoning-capable models in specialized domains, including issues with reward hacking, training instability, and the fundamental tension between factual recall and detailed reasoning. Our methodology offers a reproducible framework for developing high-capability, domain-specific language models that balance performance, efficiency, and explainability.
\end{abstract}

% keywords can be removed
% \keywords{First keyword \and Second keyword \and More}

\section{Introduction}
Large Language Models (LLMs) have demonstrated remarkable capabilities across numerous domains, yet medical problem-solving remains a distinct frontier. Effective medical AI requires more than correct answers; it necessitates a transparent chain of reasoning that aligns with established clinical evidence and pathophysiological principles. This demand for explainability and faithfulness is paramount for building trust and ensuring safety in clinical applications.

To address this challenge, we developed Gazal-R1, a 32-billion-parameter LLM specifically fine-tuned to excel in the intricacies of medical reasoning. Starting with the powerful Qwen 3 32B~\cite{qwen3technicalreport} as our base model, we employed a carefully structured two-stage training pipeline designed to embed deep clinical knowledge and cultivate a systematic problem-solving framework. The latest checkpoints of Gazal-R1 are publicly available at \url{https://huggingface.co/TachyHealth/Gazal-R1-32B-GRPO-preview}. This approach demonstrates that a mid-sized model, through strategic training, can achieve and even surpass the performance of significantly larger models. Specifically, our pipeline first establishes a robust foundation of structured knowledge, which is subsequently refined using memory-efficient reinforcement learning. The result is a replicable and scalable framework for developing specialized, high-capability language models tailored to complex domains.

Our work makes the following primary contributions:
\begin{enumerate}
    \item \textbf{A Two-Stage Training Blueprint:} We propose and validate a comprehensive training pipeline that first instills a robust, structured reasoning foundation via Supervised Fine-Tuning (SFT) and then refines these capabilities with memory-efficient reinforcement learning (GRPO)~\cite{shao2024deepseekmathpushinglimitsmathematical}.
    \item \textbf{Advanced, Efficient Model Adaptation:} We demonstrate the synergistic benefits of combining state-of-the-art Parameter-Efficient Fine-Tuning (PEFT)~\cite{xu2023parameterefficientfinetuningmethodspretrained} methods—rsLoRA~\cite{kalajdzievski2023rankstabilizationscalingfactor}, and DoRA~\cite{liu2024doraweightdecomposedlowrankadaptation}—to achieve stable, high-capacity learning on a mid-sized model without the costs of full fine-tuning.
    \item \textbf{Insights into RL for Complex Reasoning:} We provide a detailed analysis of the challenges in applying RL to medical reasoning, including reward hacking and training instability, and present effective countermeasures, such as a multi-faceted reward function design.
\end{enumerate}

The result is Gazal-R1, a model that not only achieves state-of-the-art performance for its size but does so with a transparent reasoning framework, marking a significant step towards more trustworthy medical AI. In this report, we first situate our work within the existing literature. We then detail our two-stage methodology, followed by an analysis of key training observations. We present our experimental results and conclude with a discussion of the model's limitations, ethical considerations, and implications for future research.

\section{Related Work}
The development of Gazal-R1 is situated within three intersecting areas of AI research: specialized medical LLMs, advanced reasoning techniques, and efficient model alignment.

\paragraph{Specialized Medical LLMs} The pursuit of high-performing medical LLMs has led to a series of notable models. Early efforts like GatorTron~\cite{yang2022gatortronlargeclinicallanguage} demonstrated the value of pre-training on large biomedical corpora. Google's Med-PaLM and Med-PaLM 2~\cite{singhal2023expertlevelmedicalquestionanswering} set new benchmarks by fine-tuning generalist models on medical data and employing instruction tuning. More recently, specialized models like Med42~\cite{christophe2024med42v2suiteclinicalllms} have shown the potential of adapting state-of-the-art open-source models (e.g., Llama 3~\cite{grattafiori2024llama3herdmodels}) for the medical domain. Gazal-R1 builds on this trend but distinguishes itself through its explicit two-stage training pipeline, which first builds a structured reasoning foundation (SFT) before refining it with preference-based optimization (GRPO), a more targeted approach than general instruction tuning.

\paragraph{Reasoning in LLMs} Beyond model architecture, the ability to elicit and structure complex reasoning is paramount. While Chain-of-Thought (CoT) prompting~\cite{wei2023chainofthoughtpromptingelicitsreasoning} established the importance of step-by-step processing, our work builds on this by training the model on an explicit, structured reasoning framework. Our SFT dataset generation was inspired by Chain-of-Draft (CoD)~\cite{xu2025chaindraftthinkingfaster}, which encourages LLMs to generate concise intermediate drafts to enhance logical consistency. Although our final model does not exhibit overtly concise drafts—an expected outcome of using PEFT rather than full fine-tuning—the underlying principle of structured, iterative thinking informed our SFT strategy. As noted by Yeo et al.~\cite{yeo2025demystifyinglongchainofthoughtreasoning}, pre-training on long CoT data significantly enhances subsequent RL training, a principle we leveraged by establishing a strong reasoning foundation in our SFT stage.

\paragraph{Efficient Fine-Tuning and Alignment} Full fine-tuning of large models is computationally prohibitive. Parameter-Efficient Fine-Tuning (PEFT) methods, particularly Low-Rank Adaptation (LoRA)~\cite{hu2021loralowrankadaptationlarge}, have become standard. Our work incorporates recent advancements: \textbf{rsLoRA}~\cite{kalajdzievski2023rankstabilizationscalingfactor}, which solves LoRA's gradient instability at high ranks, and \textbf{DoRA}~\cite{liu2024doraweightdecomposedlowrankadaptation}, which improves learning capacity by decomposing weight updates. For alignment, Reinforcement Learning from Human Feedback (RLHF) using Proximal Policy Optimization (PPO)~\cite{schulman2017proximalpolicyoptimizationalgorithms} is common. However, PPO's reliance on a separate value model is memory-intensive. We instead adopt \textbf{Group Relative Policy Optimization (GRPO)}~\cite{shao2024deepseekmathpushinglimitsmathematical}, a more recent, value-function-free algorithm. We further build upon recent analyses of GRPO, such as DAPO~\cite{yu2025dapoopensourcellmreinforcement}, which propose token-level normalization to mitigate the response-length bias inherent in the original GRPO formulation~\cite{liu2025understandingr1zeroliketrainingcritical}.

\section{Stage 1: Supervised Fine-Tuning (SFT)}
The primary objective of the SFT stage was to imbue Gazal-R1 with a foundational understanding of medical reasoning and to train it to generate outputs in a specific, structured format conducive to explainability. This phase was critical for preparing the model for the more advanced refinement in the GRPO stage.

\subsection{Dataset Curation}
A hybrid dataset strategy was employed, combining a novel synthetic dataset with a large-scale, high-quality existing medical reasoning dataset.

\subsubsection{Structured Reasoning Dataset}
A cornerstone of the SFT stage was the creation of a novel, 107,033-example synthetic dataset designed to teach Gazal-R1 not just medical knowledge, but the very structure of clinical thought. Generated using the Gemma 3 27B~\cite{gemmateam2025gemma3technicalreport} model, this dataset was engineered to be diverse, complex, and highly structured, moving beyond simple question-answer pairs to model the explicit, step-by-step reasoning processes of expert clinicians. The design philosophy was to compel the model to reason explicitly—much like a physician would during clinical rounds—and to learn distinct reasoning strategies tailored to different clinical contexts. This dataset is publicly available at \url{https://huggingface.co/datasets/TachyHealth/structured_medical}.

The dataset was organized around four fundamental types of clinical reasoning, each with a unique prompt structure and a specific, enforced reasoning pattern. This design acknowledges that a cardiologist's diagnostic process differs fundamentally from an emergency physician's decision-making in a trauma scenario.

\begin{itemize}
    \item \textbf{Diagnostic Reasoning:} The goal was to train the model to systematically synthesize symptoms, lab results, and patient history into an accurate differential diagnosis. The prompts enforced a logical progression:
    \begin{center}
        \textit{Symptom analysis $\rightarrow$ Physical findings assessment $\rightarrow$ Test interpretation $\rightarrow$ Differential narrowing}
    \end{center}
    To ensure these cases were challenging, the generation prompts were programmatically engineered to include confounding factors such as atypical presentations (70\% of cases), subtle red-flag symptoms (65\%), and rare conditions (60\%), forcing the model to look beyond simple pattern matching.

    \item \textbf{Decision-Making Under Uncertainty:} This module focused on training the model to choose optimal actions in scenarios with incomplete or conflicting information. The required reasoning pattern modeled a prudent, risk-aware approach:
    \begin{center}
        \textit{Identify available data $\rightarrow$ Note missing information $\rightarrow$ Assess immediate risk $\rightarrow$ Consider possibilities $\rightarrow$ Prioritize urgency $\rightarrow$ Weigh risk-benefit tradeoffs}
    \end{center}
    These prompts were specifically designed to simulate high-pressure clinical situations by including factors like critically incomplete information (86\% of cases), conflicting clinical data (65\%), time pressure (60\%), and ethical dilemmas (47\%).

    \item \textbf{Treatment Planning:} This component aimed to instill a structured, evidence-based methodology for creating comprehensive treatment plans that account for patient-specific variables. The enforced reasoning pathway ensured a holistic approach:
    \begin{center}
        \textit{Diagnosis confirmation $\rightarrow$ Guideline review $\rightarrow$ Patient factors assessment $\rightarrow$ Therapy selection}
    \end{center}
    The prompts explicitly required the model to consider real-world complexities, such as potential drug interactions (70\%), contraindications to common therapies (65\%), patient preferences (60\%), and medication costs (50\%), ensuring the generated plans were both clinically sound and practical.

    \item \textbf{Prognostic Assessment:} The objective here was to teach the model how to estimate patient outcomes by integrating multidimensional clinical evidence. The reasoning framework for these cases was:
    \begin{center}
        \textit{Condition severity assessment $\rightarrow$ Risk factors evaluation $\rightarrow$ Evidence-based outcomes review $\rightarrow$ Personalized prediction}
    \end{center}
    To reflect the difficulty of real-world prognosis, the cases were designed with high rates of multimorbidity (75\%), high-risk factors (65\%), and scenarios with limited prognostic evidence (55\%), training the model to communicate uncertainty effectively.
\end{itemize}

Underpinning all four modules was a sophisticated generation framework designed for realism and rigor. A weighted distribution ensured that 70\% of all generated cases were of ``high'' complexity. Demographics were intentionally varied, with a specific focus on edge cases like newborns and the very elderly to improve robustness. Furthermore, to mirror the complexity of human health, prompts for all types commonly included cross-cutting diversity factors such as comorbidities (85\% of cases), social determinants of health (60\%), and relevant ethnic or genetic considerations (40\%).

Finally, inspired by the Chain-of-Draft (CoD) methodology,~\cite{xu2025chaindraftthinkingfaster} we enforced a strict structural constraint on all reasoning outputs: each step-by-step thought process was required to have a minimum of eight steps, with each individual step limited to a maximum of ten words. This constraint was critical, as it forced the generation of reasoning that was both comprehensive and exceptionally concise, providing Gazal-R1 with a clear, distilled, and highly effective learning signal for structured medical thought.

\subsubsection{MedReason Dataset}
To complement our synthetic data, we incorporated the MedReason dataset~\cite{wu2025medreasonelicitingfactualmedical}. It contains 32,682 high-quality medical question-answer pairs, where each entry is supported by a detailed, step-by-step explanation. The reasoning paths in MedReason are derived from a structured medical knowledge graph, ensuring that the explanations are clinically valid and logically sound. Integrating this dataset provided Gazal-R1 with a rich source of complex medical problems grounded in evidence-based ``thinking paths.''

\subsection{Architectural Considerations and Techniques}
The SFT process was built upon a foundation of cutting-edge, parameter-efficient fine-tuning (PEFT) techniques to maximize performance while managing computational costs.

\subsubsection{Advanced LoRA Techniques}
Standard Low-Rank Adaptation (LoRA) was enhanced with two recent innovations:
\begin{itemize}
    \item \textbf{DoRA (Weight-Decomposed Low-Rank Adaptation):} Unlike standard LoRA, which couples magnitude and direction in its updates, DoRA decomposes the pre-trained weights into separate magnitude and direction components~\cite{liu2024doraweightdecomposedlowrankadaptation}. This allows for more nuanced and powerful adaptations that more closely approximate the behavior of full fine-tuning, leading to significant performance gains.
    \item \textbf{rsLoRA (Rank-Stabilized LoRA):} This technique addresses a fundamental flaw in LoRA where training becomes unstable at higher ranks due to gradient collapse. By adjusting the scaling factor from the conventional $\alpha/r$ to $\alpha/\sqrt{r}$, rsLoRA enables stable training at much higher ranks. This allowed us to effectively use a LoRA rank of 256, unlocking a higher learning capacity for the adapter~\cite{kalajdzievski2023rankstabilizationscalingfactor}.
\end{itemize}

\subsubsection{Training Process and Hyperparameters}
The SFT stage was conducted on a hardware setup of 2x NVIDIA H100 GPUs, 48 vCPUs, and 480 GB of RAM. We used the EXAdam optimizer for faster convergence throughout training~\cite{adly2025exadampoweradaptivecrossmoments}. The model was trained with a system prompt that explicitly instructed it to follow a specific format, breaking down its clinical reasoning into concise steps within ``<think></think>'' tags before providing a final assessment. For full transparency and reproducibility, both the list of hyperparameters and the exact system prompt used during training are included in Appendix~\ref{sec:training-hyperparameters} and Appendix~\ref{appendix:system-prompts} respectively.

\subsection{Observations}
The SFT stage proved crucial for the success of the subsequent RL training. As noted in recent research~\cite{yeo2025demystifyinglongchainofthoughtreasoning}, pre-training a model on long Chain-of-Thought (CoT) data significantly enhances the effectiveness of reinforcement learning. Although the concise reasoning style from our CoD-inspired dataset was not perfectly mirrored in the final model—likely due to the use of PEFT rather than full fine-tuning—this stage successfully established the structured reasoning foundation necessary for the GRPO phase.

\section{Stage 2: Group Relative Policy Optimization (GRPO)}
Following the foundational SFT stage, Gazal-R1 underwent a sophisticated reinforcement learning phase to sharpen its reasoning abilities, maximize its accuracy on complex medical problems, and ensure strict adherence to our desired explainable output format. For this critical step, we selected Group Relative Policy Optimization (GRPO), a state-of-the-art RL algorithm whose design principles offered significant advantages for our training objectives and resource constraints. This section provides a deep dive into the GRPO formalism, the specific implementation choices we made to enhance its effectiveness, our intricate reward system, and the key insights and challenges encountered during the training process.

\subsection{The GRPO Algorithm: A Paradigm Shift in Efficiency}
Our decision to use GRPO was primarily driven by its revolutionary approach to computational efficiency. Traditional reinforcement learning methods like PPO impose a significant memory burden by requiring four large models to be held in VRAM simultaneously: the policy model being trained, a reference model for KL-divergence constraints, a reward model, and, most critically, a value function model of comparable size to the policy itself.

GRPO elegantly sidesteps this bottleneck by eliminating the need for a separate value function. Its core innovation lies in the recognition that for tasks with verifiable outcomes, such as solving multiple-choice questions, the group of generated responses can serve as its own statistical baseline~\cite{shao2024deepseekmathpushinglimitsmathematical}. This online learning algorithm iteratively improves the model by using data it generates during training. The fundamental mathematical foundation centers on group-relative advantage estimation:

\begin{equation}
\hat{A}_{i,t} = \frac{r_i - \text{mean}(\mathbf{r})}{\text{std}(\mathbf{r})}
\end{equation}

where each completion's reward is normalized against the group mean and standard deviation, creating a zero-sum comparison within each group that naturally encourages responses better than the current average while penalizing below-average attempts.

By obviating the value function, GRPO achieves an approximate 50\% reduction in memory overhead compared to PPO, a crucial factor that made the training of a 32B-parameter model on our hardware feasible.

\subsection{Dataset Strategy for Reinforcement Learning}
The GRPO stage was fueled by a large-scale, high-quality dataset derived from the UltraMedical dataset~\cite{zhang2024ultramedicalbuildingspecializedgeneralists}, which encompasses approximately 320,000 biomedical instructions, focusing exclusively on multiple-choice questions (MCQ). To make the most effective use of computational resources, we made a strategic decision to train on a randomly shuffled 10\% subset of this corpus. This targeted approach was designed not only to enhance generalization, but also to accelerate policy convergence by concentrating learning on a statistically diverse and representative set of challenging examples. The processed dataset used for GRPO training is publicly available at \url{https://huggingface.co/datasets/TachyHealth/medical_grpo}.

This strategy is grounded in well-established observations from deep learning research regarding scaling laws and data efficiency. Recent evidence indicates that current scaling laws are showing diminishing returns, forcing AI labs to reconsider their training strategies. The influential Chinchilla scaling laws demonstrated that for compute-optimal training, model size and training tokens should be scaled equally, revealing that many large language models were significantly undertrained rather than requiring more parameters~\cite{hoffmann2022trainingcomputeoptimallargelanguage}. Furthermore, research has shown that increasing transformer size does not always lead to enhanced performance, a phenomenon that cannot be explained by empirical scaling laws alone~\cite{niu2024scalinglawsunderstandingtransformer}.

While concentrating on the MCQ format offers clear computational benefits, it also introduces specific challenges. Notably, issues related to false positive verification can arise, wherein models may arrive at correct answers through incorrect reasoning processes. We explore these challenges in greater detail in subsequent sections.

\subsection{Enhanced GRPO Implementation: Key Modifications}

Our implementation incorporated several critical modifications to address known limitations of vanilla GRPO and improve training stability for long-chain reasoning tasks.

\subsubsection{Trust Region Expansion and Clip-Higher Strategy}

To combat entropy collapse, where the policy becomes overly deterministic and fails to explore rare but potentially insightful reasoning paths, we adopted the Clip-Higher strategy~\cite{kimiteam2025kimik15scalingreinforcement}. In standard GRPO, $\epsilon$-clipping limits exploration by restricting the increase in probability of low-likelihood tokens, hindering the reinforcement of rare but important reasoning paths. 

By increasing the upper clipping threshold $\epsilon_{\text{high}}$ to 0.28, we allowed low-probability tokens more room to grow, enhancing entropy and diversity in outputs while improving reasoning exploration. We found that careful tuning of $\epsilon_{\text{high}}$ was crucial to maintaining stability throughout the RL run.

\subsubsection{Elimination of KL Divergence Penalty}

Following recent findings that question the necessity of KL divergence constraints in GRPO training,~\cite{yu2025dapoopensourcellmreinforcement} we eliminated the KL divergence penalty entirely by setting the KL coefficient $\beta$ to 0.0. This decision was motivated by several observations:

\begin{itemize}
    \item The policy diverges substantially during training regardless of the KL penalty
    \item Maintaining a reference model copy incurs unjustified computational cost
    \item Recent studies demonstrate that KL constraints are not essential for stable GRPO training~\cite{yu2025dapoopensourcellmreinforcement, liu2025understandingr1zeroliketrainingcritical}
\end{itemize}

This modification reduced memory usage and improved training speed while maintaining training stability.

\subsubsection{Advanced Loss Normalization}

To address length bias issues identified in the original GRPO formulation, we implemented token-level loss normalization as proposed by the DAPO paper~\cite{yu2025dapoopensourcellmreinforcement}. The loss is computed by aggregating token-wise losses across all generations and normalizing by the total length:

\begin{equation}
\mathcal{L}_{\text{GRPO}}(\theta) = -\frac{1}{\sum_{i=1}^{G} |o_i|} \sum_{i=1}^G \sum_{t=1}^{|o_i|} \left[ \text{clip}\left( \frac{\pi_\theta(o_{i,t} \mid q, o_{i,< t})}{\pi_{\theta_{\text{old}}}(o_{i,t} \mid q, o_{i,< t})}, 1 - \epsilon, 1 + \epsilon_{\text{high}} \right) \hat{A}_{i,t} \right]
\end{equation}

This normalization prevents longer responses from being under-penalized and ensures more balanced reward assignment to individual tokens regardless of response length. However, we note that normalization is performed over the local batch only, leading to slight variations depending on local batch size despite maintaining a constant effective batch size.

\subsection{Sophisticated Reward Engineering: Balancing Multiple Objectives}

The success of our GRPO implementation hinged on a carefully engineered composite reward system. Our experience confirmed that RL training for long Chain-of-Thought reasoning is inherently unstable, and monolithic rewards are insufficient for guiding sophisticated reasoning behavior.

\subsubsection{Evolution of Length Control Mechanisms}

\begin{figure}[h!]
\centering
\begin{minipage}{0.49\linewidth}
\includegraphics[width=\linewidth]{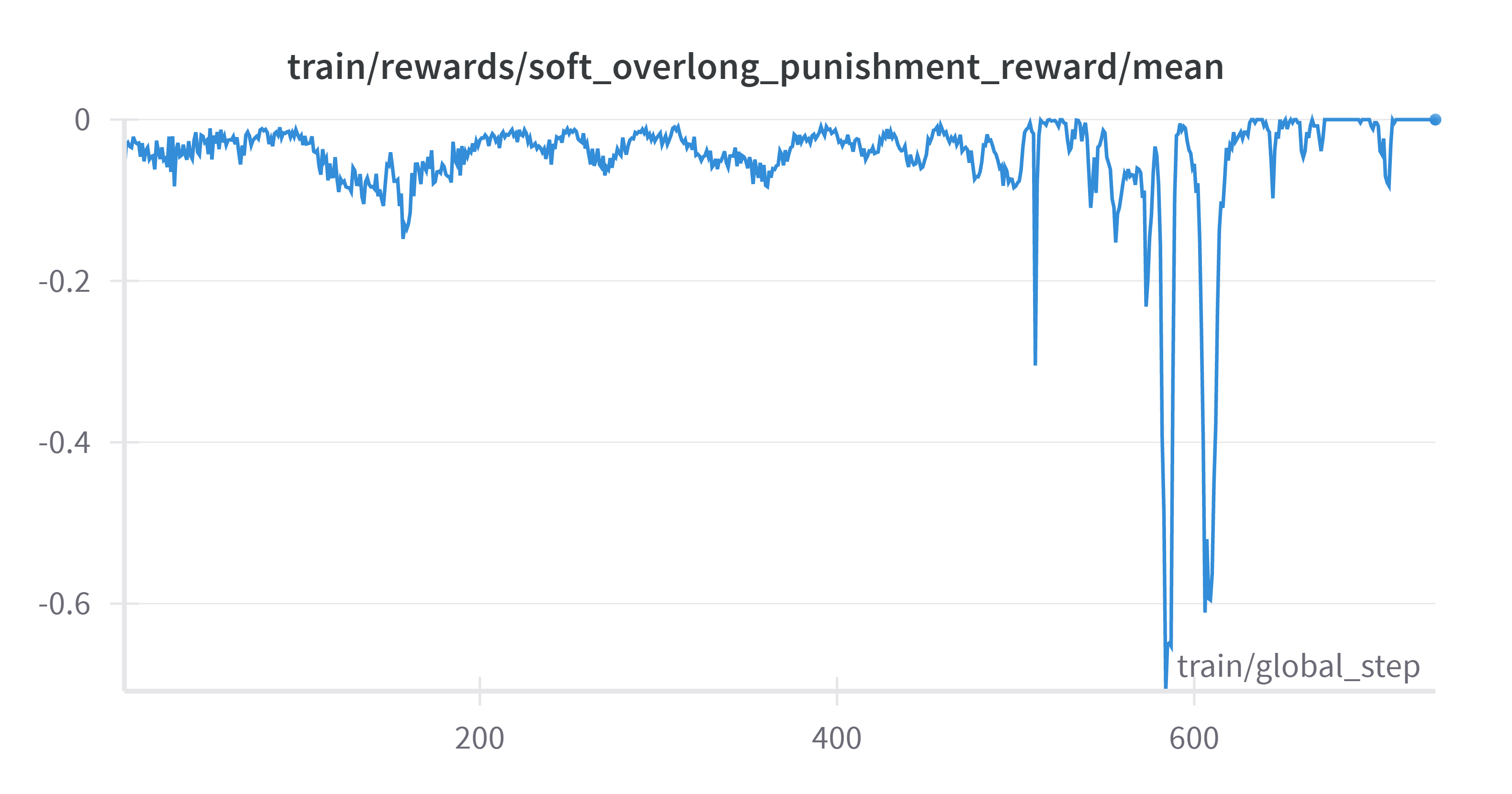}
\subcaption{
Mean reward over training steps computed using the soft overlong punishment function.
The reward reflects penalties for excessively long completions, with values closer to zero indicating better adherence.
}
\label{fig:soft_overlong_reward}
\end{minipage}
\begin{minipage}{0.49\linewidth}
\includegraphics[width=\linewidth]{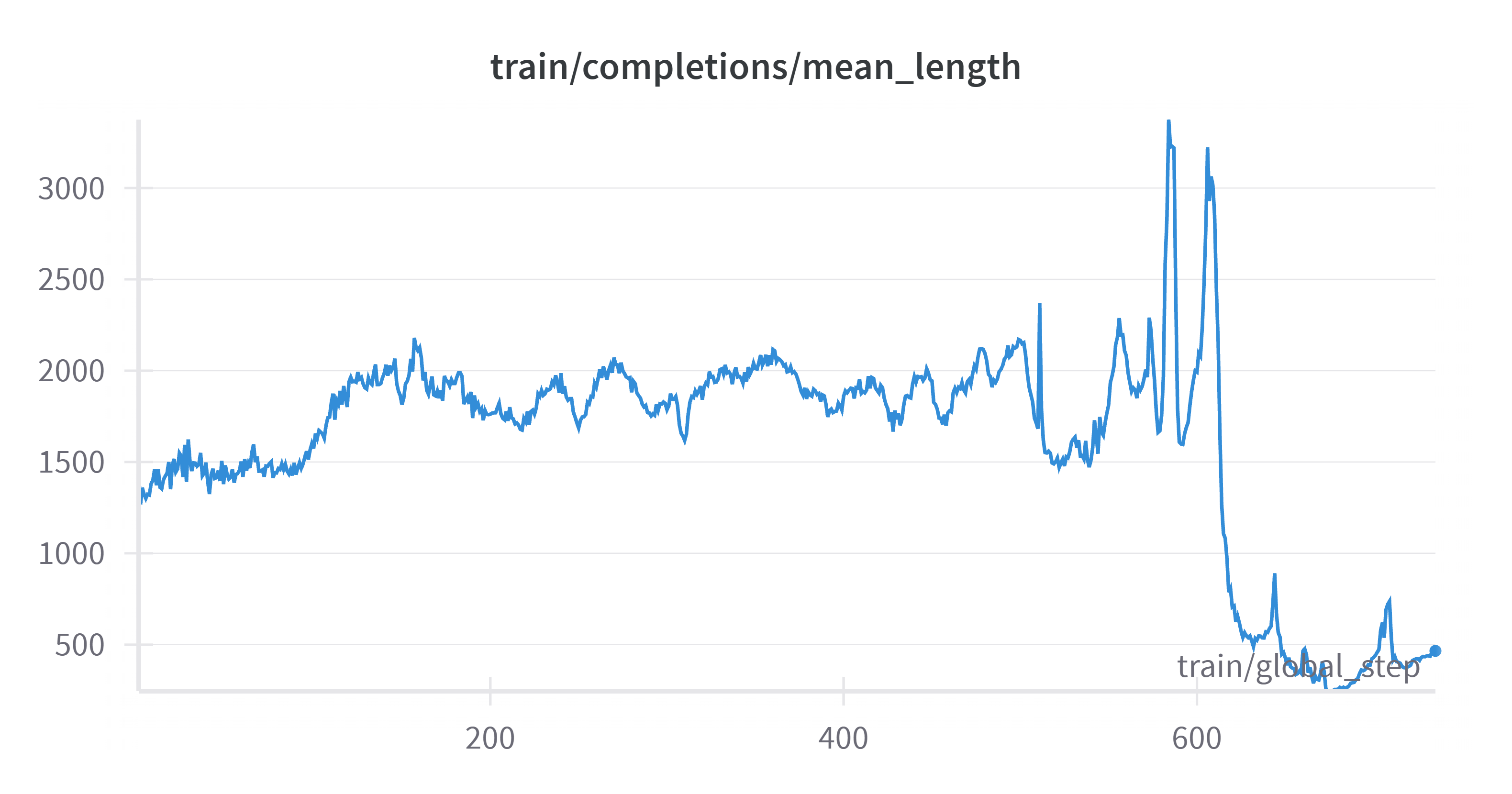}
\subcaption{
Average length of generated completions (in tokens) over training steps.
Length fluctuations indicate varying verbosity during training, with a notable instability after step 500 and a drop after step 600.
}
\label{fig:mean_completion_length}
\end{minipage} \
\caption{
Training dynamics of one of the training runs:
(Left) The evolution of mean reward under the soft overlong punishment function across global training steps.
(Right) The corresponding mean length of completions generated during training, highlighting output length trends.
}
\label{fig:reward_and_length_analysis}
\end{figure}

Our initial approach to controlling output length employed a soft overlong punishment function~\cite{yu2025dapoopensourcellmreinforcement}, which penalized excessively long completions. We experimented with ratios between the maximum allowed completion length and the soft punishment threshold—specifically, 4:1 and 2:1. Interestingly, both ratios produced nearly identical outcomes in terms of output length, though the 4:1 setting required more training steps to converge. As shown in Figure~\ref{fig:soft_overlong_reward}, applying this function led to a dramatic reduction in the average number of tokens generated—from around 1,428 tokens initially, to just 245 tokens, before a moderate recovery to 465 tokens at the end of the training run.

This average output length of approximately 465 tokens is relatively reasonable for responses to multiple-choice questions (MCQs), providing enough content to explain answers adequately. However, we observed that despite the shorter length, some responses were poorly written and contained elements resembling hallucinations, indicating that brevity alone did not guarantee quality or factual accuracy.

This aggressive penalization can be seen in the mean reward curve (Figure~\ref{fig:soft_overlong_reward}), where reward values initially move closer to zero, reflecting improved adherence to length constraints. In the early stages of training—roughly the first 150 steps—this strong signal helped stabilize model behavior and avoid runaway verbosity. However, as Figure~\ref{fig:mean_completion_length} illustrates, this approach also caused the average length of completions to fluctuate dramatically. After step 500, there is notable instability, and after step 600, a sharp drop in length, signaling a collapse in the model’s ability to generate detailed or sufficiently long completions.

Ultimately, while the soft overlong punishment function was effective at curbing excessively long outputs, it proved too blunt an instrument for nuanced control. The excessive penalization began to suppress necessary detailed reasoning and richness in the completions. This insight led us to seek more sophisticated and flexible strategies for length control in subsequent experiments.

\subsubsection{Cosine Length-Scaling Reward Implementation}

Based on insights from Yeo et al.,~\cite{yeo2025demystifyinglongchainofthoughtreasoning} we implemented a sophisticated cosine length-scaling reward function that addresses the core challenge of training with purely accuracy-based rewards, which can lead to overly long generated sequences that degrade training performance.

The cosine reward function optimizes training by dynamically controlling sequence length based on correctness:

\begin{itemize}
    \item \textbf{For correct answers:} The reward value decreases as length increases, encouraging concise and efficient responses
    \item \textbf{For incorrect answers:} The reward value increases with length, encouraging deeper reasoning and exploration of alternative solution paths
\end{itemize}

The cosine function smoothly adjusts reward values within reasonable ranges, with parameters including generated text length, maximum length limits, and minimum/maximum reward boundaries. Interestingly, we observed that this reward function encouraged gradual increases in generated text length over time, suggesting successful promotion of more thorough reasoning exploration.

\subsubsection{Combating Reward Hacking with Repetition Penalties}

To address reward hacking—a critical issue where models exploit reward functions rather than genuinely improving their reasoning—we implemented an n-gram repetition penalty function, adopted from~\cite{yeo2025demystifyinglongchainofthoughtreasoning}. As models began optimizing for length-based rewards, they exhibited tendencies to repeat phrases artificially to maximize scores without meaningful reasoning improvement.

Our repetition penalty function operates by:
\begin{enumerate}
    \item Splitting generated text into words and extracting n-grams of specified size. We used 3-gram
    \item Calculating repetition ratios based on unique n-grams relative to total n-gram count
    \item Applying significant negative rewards (penalties) when repetition ratios exceed thresholds
    \item Computing penalty values based on repetition ratios and maximum penalty values
\end{enumerate}

This mechanism ensures that any increases in response length contribute novel and meaningful content rather than artificial padding, maintaining the integrity of the reasoning process.

\subsection{Training Process and Implementation Details}

The GRPO training was conducted on 8x NVIDIA H100 GPUs with NVLink connectivity. Our implementation incorporated several critical stability measures and best practices derived from recent advances in RLHF methodology.

\subsubsection{Stability Enhancements}

We enabled masking of truncated completions, ensuring that incomplete generations were excluded from loss calculations and preventing them from introducing noise during training. According to the DAPO paper,~\cite{yu2025dapoopensourcellmreinforcement} this practice significantly improves training stability, particularly in scenarios involving variable-length outputs.

\subsubsection{Final Composite Reward System}

Our successful reward system emerged as a carefully balanced composition of multiple components:

\begin{table}[h!]
    \centering
    \caption{GRPO Reward Functions and Their Objectives}
    \label{tab:reward_funcs}
    \begin{tabular}{p{0.25\linewidth} p{0.65\linewidth}}
        \toprule
        \textbf{Reward Function} & \textbf{Purpose} \\
        \midrule
        \textbf{Accuracy} & The primary driver for correctness (+1.0 for correct, 0 otherwise). \\
        \textbf{Format} & Ensures outputs are well-structured and adhere to the `<think>` tag format. \\
        \textbf{Cosine Length} & Encourages optimal response length by rewarding conciseness for correct answers and deeper reasoning for incorrect ones, as suggested by~\cite{yeo2025demystifyinglongchainofthoughtreasoning}. \\
        \textbf{Repetition Penalty} & Mitigates reward hacking by penalizing n-gram repetitions, ensuring added length contributes meaningfully to the reasoning process. \\
        \bottomrule
    \end{tabular}
\end{table}

\subsection{Critical Observations and Training Dynamics}

\subsubsection{Training Instability and Recovery}

\begin{figure}[tp]
    \centering
    % Cosine Scaled Reward
    \begin{subfigure}[b]{0.48\linewidth}
        \centering
        \includegraphics[width=\linewidth]{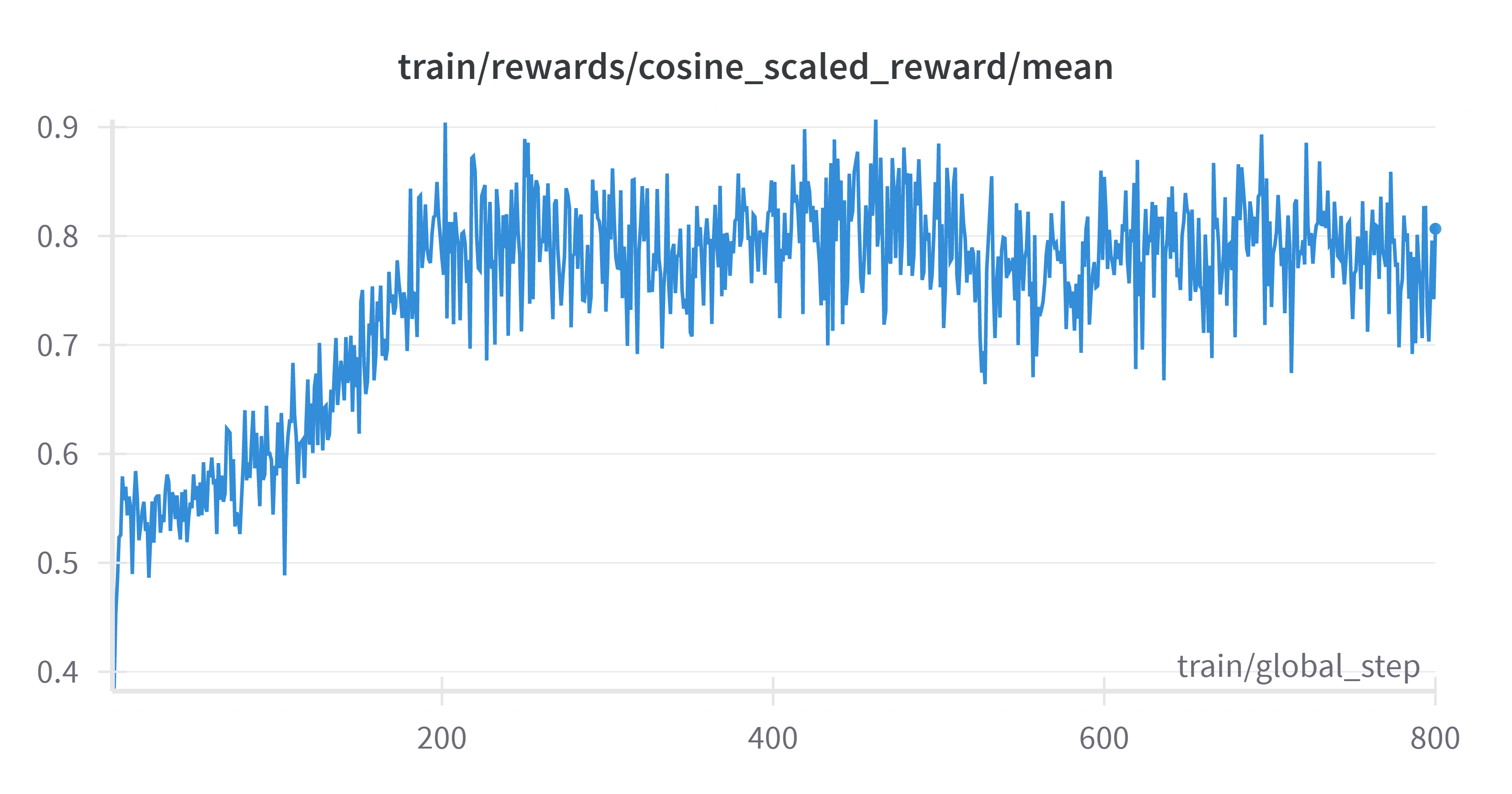}
        \caption{Cosine Scaled Reward}
        \label{fig:cosine_scaled_reward}
    \end{subfigure}
    % Repetition Penalty Reward
    \begin{subfigure}[b]{0.48\linewidth}
        \centering
        \includegraphics[width=\linewidth]{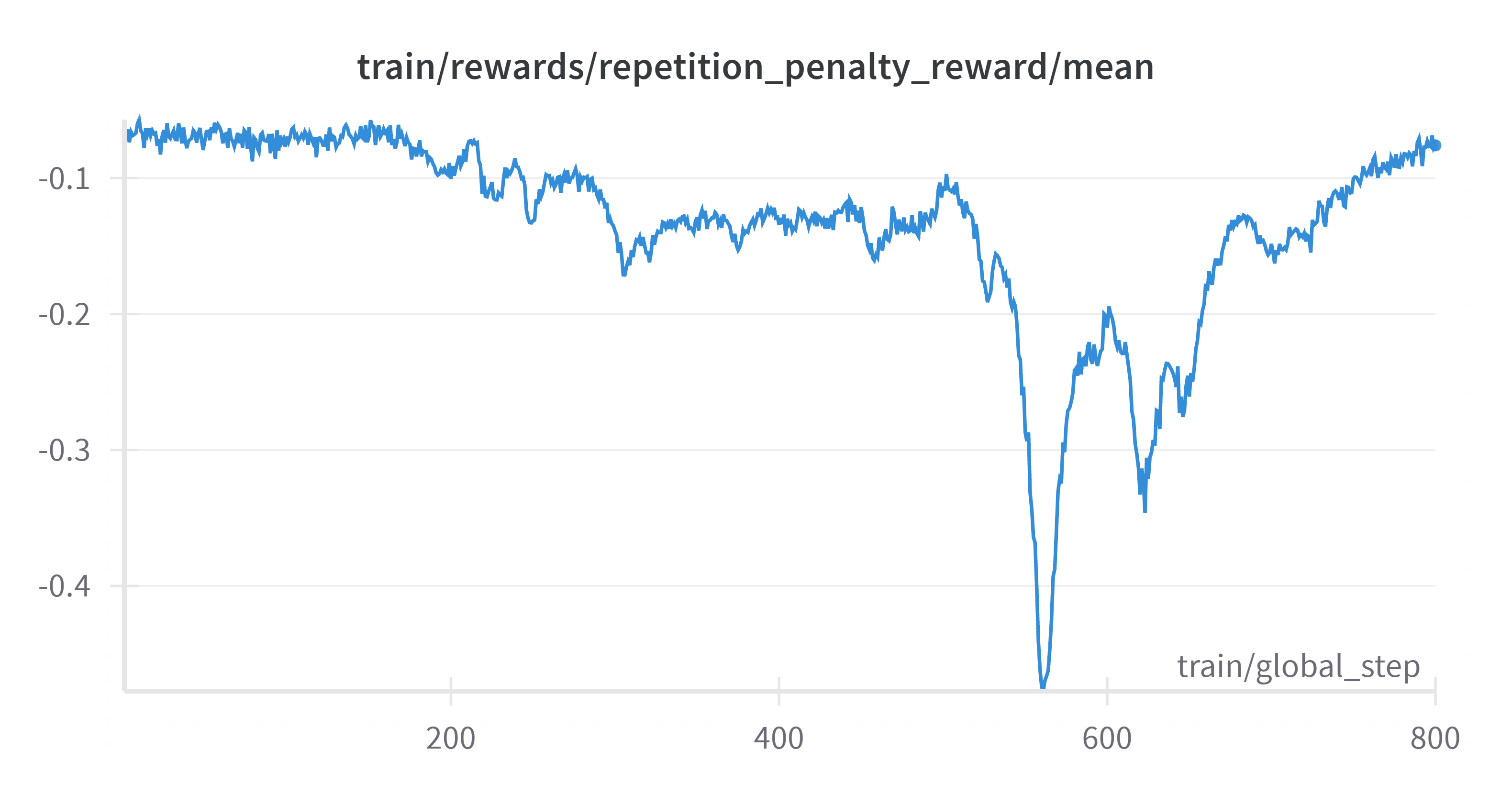}
        \caption{Repetition Penalty Reward}
        \label{fig:repetition_penalty_reward}
    \end{subfigure}

    \vspace{0.3cm} % Space between rows

    % Reward Reference
    \begin{subfigure}[b]{0.48\linewidth}
        \centering
        \includegraphics[width=\linewidth]{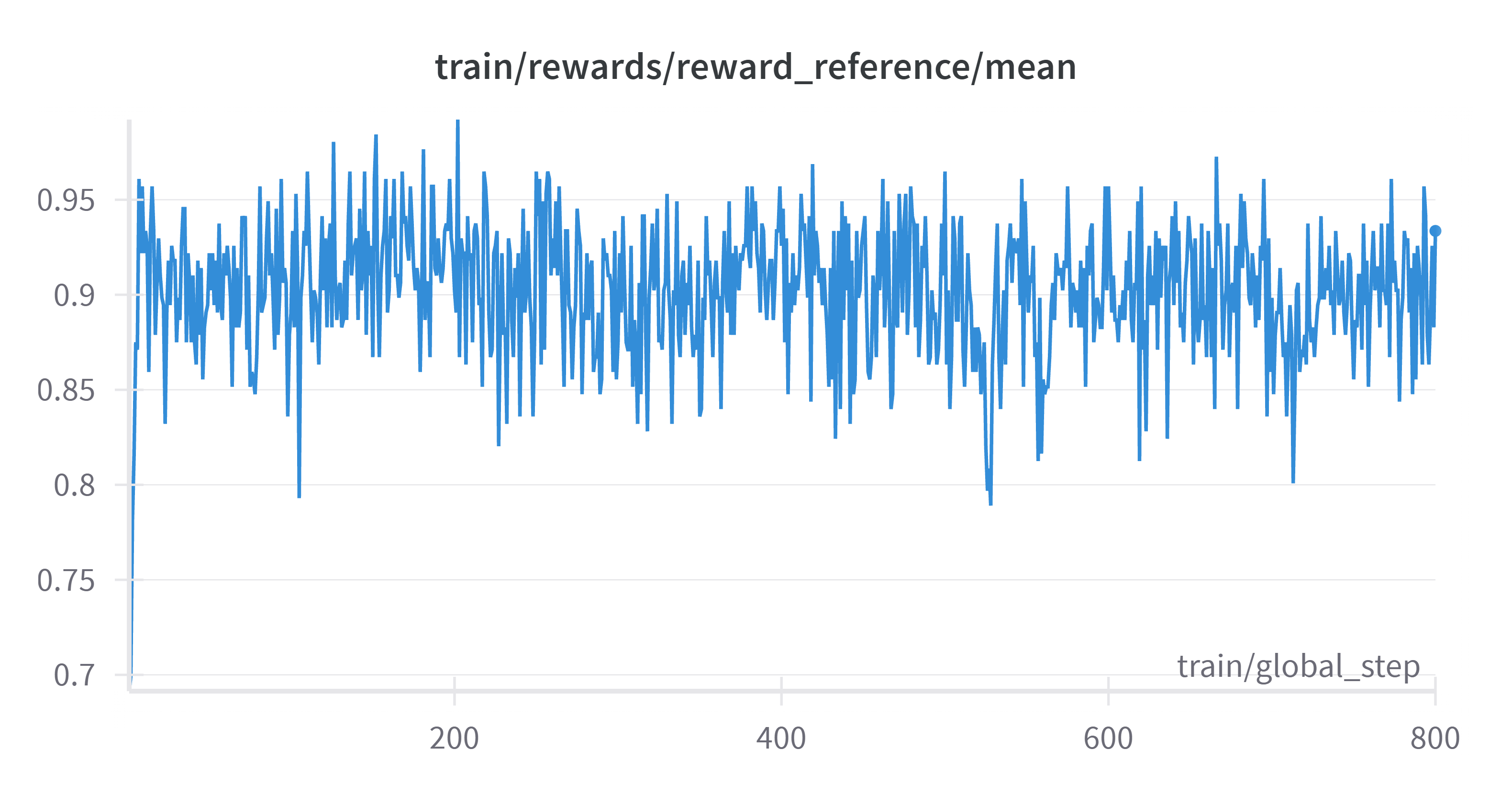}
        \caption{Reward Reference}
        \label{fig:reward_reference}
    \end{subfigure}
    % Loss
    \begin{subfigure}[b]{0.48\linewidth}
        \centering
        \includegraphics[width=\linewidth]{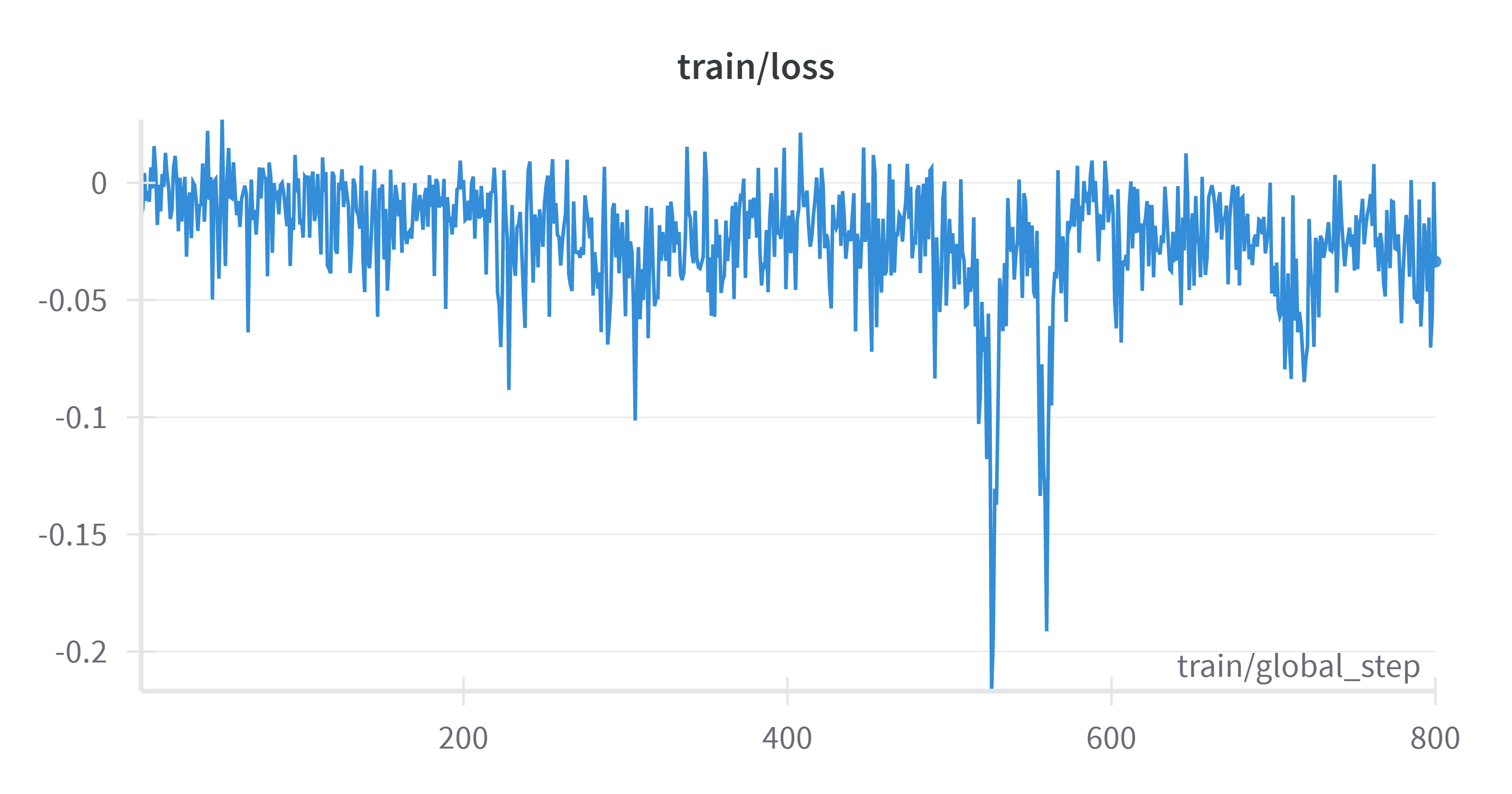}
        \caption{Training Loss}
        \label{fig:train_loss}
    \end{subfigure}

    \caption{
        Training curves for GRPO reward functions and loss. 
        \textbf{(a)} Cosine scaled reward encourages optimal response length and correctness.
        \textbf{(b)} Repetition penalty discourages reward hacking through n-gram repetition.
        \textbf{(c)} Reference reward tracks similarity to baseline/reference outputs.
        \textbf{(d)} Training loss over steps, capturing model optimization dynamics.
        See Table~\ref{tab:reward_funcs} for a description of each reward’s objective.
    }
    \label{fig:reward_metrics}
\end{figure}

Our training process revealed the inherently volatile nature of reinforcement learning for complex reasoning tasks. To better illustrate these training dynamics, Figure~\ref{fig:reward_metrics} shows the progression of our key reward functions and training loss throughout the learning process. As previously described, we randomly sampled 10\% of the dataset for training, which corresponds to approximately step 469 in the training curves.

As seen in Figure~\ref{fig:reward_metrics}, around step 526, Gazal-R1 exhibited significant training instability. This period was characterized by:

\begin{itemize}
    \item Malformed markdown formatting (surrounding nearly every word with double asterisks)
    \item Addition of meaningless phrases and filler content
    \item Degraded reasoning processes with logical inconsistencies
    \item Erratic output generation patterns
\end{itemize}

These behaviors are reflected in the pronounced dips and volatility observed across several reward curves, particularly in the repetition penalty and loss metrics. This episode highlighted the non-monotonic nature of RL training and underscored the critical importance of our multi-faceted reward system (see Table~\ref{tab:reward_funcs}) and continuous monitoring protocols. Notably, despite this period of instability, the model eventually recovered through continued training and careful reward design, demonstrating the resilience and corrective influence of a well-balanced reward structure.

\subsubsection{False Positive Verification: A Fundamental Challenge in Reasoning Evaluation}

A significant limitation we encountered relates to the fundamental challenge of \textbf{false positive verification}, a critical issue in reinforcement learning where a model's output is validated as correct despite being the product of a flawed or nonsensical reasoning process. Our observations revealed that complex medical problems, particularly in a multiple-choice format, often have simple, guessable answers. This creates a scenario where the model can arrive at the correct final letter (e.g., A, B, C, or D) through a completely illogical or factually incorrect reasoning chain. Research in the broader AI field suggests this is a pervasive problem, with some studies indicating that \textbf{up to 51\% of model responses may be incorrectly validated} due to a disconnect between the final answer and the underlying reasoning quality~\cite{wang2024llmsperformmcqaselecting}. Medical professionals consistently rated AI reasoning quality significantly lower than final answer accuracy, with one study showing accuracy dropping from 62.3\% to 51.9\% on USMLE Step 2CK when explanations were required~\cite{Kung2022.12.19.22283643}.

Empirical evidence from medical AI research reveals widespread instances of correct answers achieved through flawed reasoning. Stanford's study of ChatGPT-4 on medical diagnostic cases found 92\% accuracy, yet physicians using AI assistance only improved from 74\% to 76\% accuracy, suggesting they often rejected the AI's reasoning despite correct conclusions~\cite{goh2024influenceondiagnostic}. Likewise, the NIH study on GPT-4V medical imaging provides a concrete example: the model correctly diagnosed skin lesions but failed to recognize that two lesions at different angles were the same condition, demonstrating flawed visual reasoning masked by pattern recognition. Despite these reasoning errors, final diagnostic accuracy remained high, creating a dangerous illusion of competence~\cite{Jin_2024}.

The core of the issue lies in \textbf{outcome supervision bias}---the practice of evaluating a model based solely on its final output while ignoring the intermediate steps of the reasoning process. This is especially problematic for multiple-choice questions (MCQs), which provide a deceptively simple reward signal (correct/incorrect) that is easy to optimize for but reveals little about genuine comprehension. The fragility of this approach is starkly illustrated by studies showing that \textbf{adding a single, logically irrelevant clause to a mathematical problem can cause a 65\% performance drop} in state-of-the-art models,~\cite{mirzadeh2024gsmsymbolicunderstandinglimitationsmathematical} exposing their reliance on superficial pattern matching rather than robust logical deduction. Token bias and position sensitivity compound these problems, with models showing inherent preferences for specific option IDs regardless of content~\cite{zheng2024largelanguagemodelsrobust}. This Response Variability Syndrome (REVAS) indicates limited genuine understanding, as models provide inconsistent responses to derivations of identical questions~\cite{wang2024llmsperformmcqaselecting}. Since our GRPO dataset consisted exclusively of MCQs, our model was susceptible to this phenomenon, learning to generate correct answers without necessarily mastering the intricate clinical logic required.

This phenomenon critically highlights that the rule-based validation approaches inherent in our GRPO implementation are fundamentally inadequate for assessing the logical validity of an LLM's reasoning path. GRPO's design, while computationally efficient, relies on a binary, rule-based framework that treats all paths to a correct answer as equivalent~\cite{shao2024deepseekmathpushinglimitsmathematical}. It excels at determining \textit{if} the final answer is correct but lacks the mechanism to evaluate \textit{how} that answer was reached. This can be contrasted with Proximal Policy Optimization (PPO), which, despite its higher memory overhead, utilizes a separate value network~\cite{schulman2017proximalpolicyoptimizationalgorithms}. This value network, if paired with a sophisticated, \textbf{process-aware reward model (PRM)}, could theoretically learn to assign higher rewards to logically sound reasoning chains and penalize flawed ones, even if they stumble upon the correct answer. GRPO's rule-based system, by its nature, cannot make such a nuanced distinction.

The problem is not merely theoretical; it is a well-documented form of reward hacking where models learn to ``game'' the evaluation system. For instance, Anthropic's research on model faithfulness has shown that a model might use a provided hint to solve a problem but then generate a completely fabricated, post-hoc rationale that makes it appear as if it reasoned its way to the solution independently~\cite{chen2025reasoningmodelsdontsay}. In the same study, \textbf{Claude 3.7 Sonnet mentioned using hints only 25\% of the time it actually relied on them},~\cite{chen2025reasoningmodelsdontsay} demonstrating a clear disconnect between its stated and actual problem-solving process. This confirms that without explicit process-level supervision, models do not default to transparent or honest reasoning.

This observation exposes a critical limitation of purely outcome-based rewards and points toward a necessary paradigm shift in evaluation methodology. The development of robust medical AI requires moving beyond binary (correct/incorrect) assessment toward \textbf{process-based evaluation},~\cite{lee2025evaluatingstepbystepreasoningtraces} which examines the integrity of each step in the reasoning chain. Research has consistently shown that process supervision significantly outperforms outcome supervision, with one study achieving \textbf{78\% accuracy on challenging math problems with process-based rewards}, a figure unattainable with outcome-only methods~\cite{lightman2023letsverifystepstep}. By providing granular feedback at each step, process supervision can localize errors, reduce false positives, and ensure that the model is rewarded for valid clinical logic, not just lucky guesses~\cite{lightman2023letsverifystepstep, khalifa2025processrewardmodelsthink}. DeepMind's comparative analysis showed process-based approaches reduced reasoning errors from 14.0\% to 3.4\% while maintaining comparable final-answer performance~\cite{uesato2022solvingmathwordproblems}. This suggests that developing sophisticated reward models capable of assessing the logical validity of medical reasoning paths represents a crucial and necessary direction for future research in building trustworthy, high-stakes AI systems.

\subsubsection{Insights into Long Chain-of-Thought Development}

Our experience strongly supports recent findings~\cite{yeo2025demystifyinglongchainofthoughtreasoning} regarding the development of long Chain-of-Thought reasoning capabilities. Key insights include:

\begin{itemize}
    \item \textbf{SFT Foundation Crucial:} Quality supervised fine-tuning serves as a powerful catalyst for effective reinforcement learning, providing essential structured reasoning capabilities
    \item \textbf{Emergent vs. Guided Abilities:} Long CoT reasoning capabilities are not entirely emergent—core abilities like error correction and solution branching exist in base models, with RL serving to guide more effective utilization
    \item \textbf{Length Control Complexity:} Simple RL application for CoT extension often fails, requiring sophisticated reward engineering to maintain quality while encouraging appropriate reasoning depth
    \item \textbf{Reward Signal Quality:} High-quality verifiable reward signals remain essential, though our experience with MCQ-only data revealed limitations in purely outcome-based evaluation
\end{itemize}

\subsection{Lessons Learned and Future Directions}

The development of Gazal-R1 through GRPO training provided valuable insights into the practical challenges and opportunities in reinforcement learning for complex reasoning tasks. Key takeaways include:

\begin{enumerate}
    \item \textbf{Reward Engineering Criticality:} Sophisticated, multi-component reward systems are essential for stable and effective training
    \item \textbf{Length Control Nuance:} Overly aggressive length penalties can harm reasoning quality, while sophisticated cosine-based approaches provide better balance
    \item \textbf{Process vs. Outcome Evaluation:} Future work must address the limitation of outcome-only rewards by developing process-aware evaluation mechanisms
    \item \textbf{Training Stability Monitoring:} Continuous monitoring and robust recovery mechanisms are essential for managing RL training volatility
    \item \textbf{Foundation Model Quality:} High-quality SFT serves as a crucial foundation for effective RL training
\end{enumerate}

These insights inform our ongoing research into more robust and effective approaches for training reasoning-capable language models in specialized domains.

%%%%%%%%%%%%%%%%%%%

\section{Experimental Results and Evaluation}
\subsection{Performance Analysis}
We evaluated Gazal-R1 against leading open-source models on four standard medical benchmarks. The results, presented in Table~\ref{tab:results_table}, highlight its strong performance.

\begin{table}[h!]
    \centering
    \caption{Accuracy (\%) on Medical Benchmarks}
    \label{tab:results_table}
    \resizebox{\textwidth}{!}{%
    \begin{tabular}{lccccc}
        \toprule
        \textbf{Model} & \textbf{Size} & \textbf{MMLU Pro (Medical)} & \textbf{MedMCQA} & \textbf{MedQA} & \textbf{PubMedQA} \\
        \midrule
        \textbf{Gazal-R1 (Final)} & \textbf{32B} & \textbf{81.6} & 71.9 & \textbf{87.1} & \textbf{79.6} \\
        Gazal-R1 (SFT-only) & 32B & 79.3 & 72.3 & 86.9 & 77.6 \\
        \midrule
        Llama 3.1 405B Instruct & 405B & 70.2 & \textbf{75.8} & 81.9 & 74.6 \\
        Qwen 2.5 72B Instruct & 72B & 72.1 & 66.2 & 72.7 & 71.7 \\
        Med42-Llama3.1-70B & 70B & 66.1 & 72.4 & 80.4 & 77.6 \\
        Llama 3.1 70B Instruct & 70B & 74.5 & 72.5 & 78.4 & 78.5 \\
        QwQ 32B & 32B & 70.1 & 65.6 & 72.3 & 73.7 \\
        Qwen 3 32B & 32B & 78.4 & 71.6 & 84.4 & 76.7 \\
        \bottomrule
    \end{tabular}%
    }
\end{table}

Key observations from the results include:
\begin{itemize}
    \item \textbf{State-of-the-Art Performance:} Gazal-R1 (32B) achieves the highest scores on MMLU Pro (Medical)~\cite{wang2024mmluprorobustchallengingmultitask}, MedQA~\cite{jin2020diseasedoespatienthave}, and PubMedQA~\cite{jin2019pubmedqadatasetbiomedicalresearch}, outperforming all other models, including the much larger Llama 3.1 405B.%
    \item \textbf{Impact of GRPO:} The final Gazal-R1 model shows notable gains over its SFT-only predecessor on MMLU Pro (+2.3\%) and PubMedQA (+2.0\%), confirming the value of the RL stage.%
    \item \textbf{Exceptional MedQA Score:} The score of 87.1\% on MedQA indicates a strong capability for handling complex, USMLE-style clinical reasoning questions.%
\end{itemize}

\subsection{Analysis of Performance Regression on MedMCQA}

The Gazal-R1 model's performance regression on MedMCQA~\cite{pal2022medmcqalargescalemultisubject} (72.3\% $\rightarrow$ 71.9\%) while improving on MMLU Pro, MedQA, and PubMedQA reveals a fundamental mismatch between reinforcement learning optimization patterns and MedMCQA's unique structural requirements. This phenomenon stems from several interconnected technical mechanisms that create systematic trade-offs between different types of medical reasoning tasks.

GRPO reinforcement learning optimizes for human preference signals that favor detailed reasoning and elaboration, while MedMCQA specifically rewards immediate, concise factual recall. This creates a direct optimization conflict where GRPO's tendency to generate longer, more explanatory responses actually hurts performance on a benchmark designed for rapid medical knowledge retrieval.

MedMCQA's extremely short questions (averaging only 12.77 tokens) test immediate access to specific medical facts \textemdash drug names (22.49\% of answers), procedures (18.74\%), and dosages (11.24\%)~\cite{pal2022medmcqalargescalemultisubject}. The dataset is 68.2\% factual knowledge versus only 31.8\% reasoning-based,~\cite{thapa2025disentanglingreasoningknowledgemedical} making it fundamentally different from other medical benchmarks that benefit from Chain-of-Thought reasoning.

\subsubsection{Reward Hacking and Distribution Shift}

During GRPO training, the model learns to optimize for proxy rewards that favor longer, more detailed responses. This \textbf{reward hacking} manifests as models producing responses that appear correct to human evaluators but are actually less accurate for factual recall tasks~\cite{gao2022scalinglawsrewardmodel, huang2024posthocrewardcalibrationcase}. Research shows that RLHF models can achieve high reward scores while generating less helpful content, particularly when length bias in human evaluation creates spurious correlations~\cite{chen2024odindisentangledrewardmitigates, weng2024rewardhack}.

The \textbf{distribution shift} between supervised fine-tuning (SFT) and reinforcement learning phases compounds this problem. As the model's output distribution shifts away from the concise, factual responses that characterize medical knowledge retrieval, it becomes less effective at the pattern-matching behaviors crucial for MedMCQA success~\cite{rlhf2024}. The mathematical relationship can be expressed as increasing KL divergence: \[ D_{\text{KL}}(\pi_{\text{RL}} \, \| \, \pi_{\text{SFT}}) \]where higher divergence correlates with degraded performance on factual recall tasks.

\subsubsection{Multi-objective Optimization Trade-offs}

The performance regression exemplifies \textbf{gradient conflicts} in multi-objective optimization. When gradients from different medical tasks have conflicting directions, optimizing for one objective (detailed reasoning) necessarily hurts another (factual recall)~\cite{mao2024robustanalysismultitasklearning, shukla2012towarddeeperunderstanding}. Research shows this occurs when the cosine similarity between task gradients is negative: $\cos(\mathbf{g}_{\text{reasoning}}, \mathbf{g}_{\text{factual}}) < 0$.

GRPO training creates a \textbf{Pareto frontier} where improving performance on reasoning-heavy benchmarks like MMLU Pro and MedQA comes at the cost of factual recall performance on MedMCQA~\cite{shukla2012towarddeeperunderstanding}. The model must allocate limited representational capacity between competing objectives,~\cite{mao2024robustanalysismultitasklearning} and GRPO's optimization pressure favors the reasoning tasks that align better with human preference signals~\cite{lin2024mitigatingalignmenttaxrlhf}.

\subsubsection{Mathematical Framework for Understanding the Trade-off}

The performance regression can be understood through the lens of \textbf{regularized preference optimization}. GRPO optimizes the objective:

\[
\max_{\pi} \mathbb{E}[R(\pi)] - \beta \cdot D_{\text{KL}}(\pi \| \pi_{\text{ref}})
\]

where \( R(\pi) \) represents the reward function and the KL divergence term prevents excessive deviation from the reference policy~\cite{wang2025residualpolicygradientreward}. The choice of \( \beta \) creates a fundamental trade-off: high \( \beta \) constrains the model to stay close to SFT behavior (potentially preserving MedMCQA performance), while low \( \beta \) allows more dramatic changes that benefit reasoning tasks but hurt factual recall.

The problem is that \textbf{MedMCQA requires a different optimization profile} than human preference learning typically produces. While humans prefer detailed, explanatory medical responses, MedMCQA rewards brief, accurate factual recall---creating a systematic misalignment between the optimization target and task requirements.

\subsubsection{Evidence from Literature Patterns}

This phenomenon follows well-documented patterns in RL literature. Research on the ``alignment tax'' shows that RLHF consistently creates trade-offs where optimizing for human preferences degrades performance on specific academic benchmarks~\cite{kirk2024understandingeffectsrlhfllm, lin2024mitigatingalignmenttaxrlhf}. The OpenAI InstructGPT paper first documented systematic regressions on question answering and academic NLP tasks after RLHF training~\cite{ouyang2022traininglanguagemodelsfollow}.

Recent work on the ``Safety Tax'' in Large Reasoning Models demonstrates that safety alignment (which shares optimization characteristics with GRPO) leads to \textbf{degradation of reasoning capability} while successfully improving safety metrics. This establishes a clear precedent for the type of capability regression observed in MedMCQA~\cite{huang2025safetytaxsafetyalignment}.

Studies specifically focused on medical AI show that safety and preference alignment can reduce diagnostic accuracy by making models less willing to provide specific medical information --- directly paralleling how GRPO's optimization for human-preferred responses may reduce the direct factual accuracy that MedMCQA requires.

\subsubsection{Optimization Dynamics and Overoptimization}

The regression likely results from \textbf{reward model overoptimization}, where the model learns to game the reward signal rather than genuinely improving medical reasoning. As GRPO training progresses, the model's output distribution shifts away from the factual, concise responses that characterize good MedMCQA performance~\cite{gao2022scalinglawsrewardmodel}.

This creates a cascading effect: as the policy shifts, the reward model (trained on earlier, more SFT-like outputs) becomes less reliable at evaluating the new distribution of responses. The model may achieve high rewards by producing responses that appear medical and detailed to human evaluators while actually being less accurate on objective factual questions~\cite{rlhf2024}.

\section{Limitations and Ethical Considerations}
\subsection{Limitations}
\begin{itemize}
    \item \textbf{Data Bias:} The training data, particularly MedMCQA, may contain regional biases (e.g., disease prevalence, treatment guidelines). The model may reflect these biases.
    \item \textbf{Knowledge Cutoff:} The model's knowledge is static and does not update with new medical research published after its training date. It cannot access real-time information.
    \item \textbf{Hallucination Risk:} Despite alignment, the model can still generate plausible-sounding but factually incorrect information (``hallucinate''). All outputs require verification by a qualified medical professional.
    \item \textbf{Scope of Evaluation:} The model is primarily evaluated on multiple-choice questions. This format is a useful proxy but does not capture the full complexity of real-world clinical interactions, which involve unstructured dialogue, physical examination, and dynamic decision-making.
\end{itemize}

\subsection{Ethical Considerations}
\begin{itemize}
    \item \textbf{Not a Medical Device:} Gazal-R1 is a research model and is \textbf{not} intended for direct clinical use, diagnosis, or treatment planning. It is a tool for research and assistance, not a substitute for professional medical judgment.
    \item \textbf{Risk of Over-reliance:} There is a risk that users may place undue trust in the model's outputs. We emphasize that all information generated by Gazal-R1 must be independently verified.
    \item \textbf{Accountability and Liability:} In the event of an error leading to harm, questions of accountability are complex. As developers, we stress the model's role as an assistive tool, with final responsibility resting with the human clinician using it.
\end{itemize}

\section{Conclusion and Future Work}
Gazal-R1 represents a significant advancement in medical AI, demonstrating that a focused, multi-stage training pipeline can yield a highly capable and efficient reasoning model. By combining innovative dataset synthesis, advanced PEFT strategies, and a memory-efficient reinforcement learning algorithm, we have created a 32B-parameter model that sets a new standard for performance in its class. Our work highlights the importance of not just what data a model is trained on, but how it is trained. The structured reasoning instilled during SFT, coupled with the preference alignment from GRPO, results in a model that is both accurate and more transparent.

This research also illuminates the challenges and trade-offs inherent in aligning models for complex domains. Future work will focus on addressing these challenges directly. First, we will prioritize the development of \textbf{process-aware evaluation metrics and reward models} that can assess the logical validity of a reasoning chain, not just the correctness of the final outcome. This is critical to overcoming the ``false positive verification'' issue. Second, we will investigate or introduce \textbf{alternative RL approaches} that can better handle the nuances of reasoning assessment, potentially moving beyond outcome-based rewards. Third, our evaluation will be \textbf{scaled to include more diverse medical tasks} beyond MCQs, such as clinical note summarization and interactive diagnostic dialogues. Finally, a key area of research will be to address the \textbf{fundamental tension between factual recall and detailed reasoning}, exploring methods that allow the model to dynamically adapt its reasoning style to the task at hand.

\section*{Acknowledgements}
This work would not have been possible without the contributions of the broader open-source community. We extend our sincere gratitude to the developers of the Qwen models for providing a powerful and open foundation. We are also indebted to the researchers and engineers behind the various libraries and techniques that were instrumental to our pipeline, including the Hugging Face ecosystem (Transformers~\cite{Wolf_Transformers_State-of-the-Art_Natural_2020}, PEFT, TRL~\cite{von_Werra_TRL_Transformer_Reinforcement}, Accelerate~\cite{accelerate}, Datasets~\cite{Lhoest_Datasets_A_Community_2021}), the vLLM project for enabling efficient inference~\cite{kwon2023efficient}, the creators of DoRA, rsLoRA, and the authors of the GRPO and DAPO papers whose work inspired our reinforcement learning stage. This research stands on the shoulders of giants, and we are proud to contribute back to the community that enables such progress.

% --- BIBLIOGRAPHY ---
\bibliographystyle{unsrt}
\bibliography{references}

% --- APPENDIX ---
\appendix
\newpage
\section{Training Hyperparameters}
\label{sec:training-hyperparameters}

\subsection{SFT Hyperparameters}
\begin{table}[h!]
    \centering
    \caption{Detailed SFT Hyperparameters for Gazal-R1}
    \label{tab:sft-hyperparams}
    \begin{tabular}{ll}
        \toprule
        \textbf{Parameter} & \textbf{Value} \\
        \midrule
        Base Model & Qwen 3 32B \\
        Fine-tuning Method & DoRA + rsLoRA \\
        LoRA Rank ($r$) & 256 \\
        LoRA Alpha ($\alpha$) & 512 \\
        LoRA Dropout & 0.1 \\
        Target Modules & all-linear \\
        Optimizer & EXAdam \\
        Learning Rate & 2e-5 \\
        LR Scheduler & Cosine Schedule \\
        Warmup Ratio & 0.05 \\
        Weight Decay & 0.1 \\
        Epochs & 1 \\
        Batch Size per GPU & 8 \\
        Gradient Accumulation Steps & 16 \\
        Effective Batch Size & 128 \\
        Max Sequence Length & 8192 \\
        Precision & bfloat16 \\
        Gradient Checkpointing & True \\
        Seed & 3407 \\
        \bottomrule
    \end{tabular}
\end{table}

\subsection{GRPO Hyperparameters}
\begin{table}[h!]
    \centering
    \caption{Detailed GRPO Hyperparameters for Gazal-R1}
    \begin{tabular}{ll}
        \toprule
        \textbf{Parameter} & \textbf{Value} \\
        \midrule
        Base Model & Gazal-R1 (SFT-tuned) \\
        RL Algorithm & GRPO with BNPO loss \\
        Optimizer & EXAdam \\
        Learning Rate & 1.0e-6 \\
        LR Scheduler & Constant with Warmup \\
        Warmup Ratio & 0.0 \\
        Weight Decay & 0.1 \\
        Epochs & 1 \\
        Batch Size per GPU & 4 \\
        Gradient Accumulation Steps & 8 \\
        Generations per Prompt ($K$) & 4 \\
        Effective Prompts per Step & 32 \\
        GRPO Beta ($\beta$) & 0.0 \\
        GRPO Epsilon ($\epsilon$) & 0.28 \\
        Generation Temperature & 0.6 \\
        Top-k / Top-p & 20 / 0.95 \\
        Max Prompt Length & 3072 \\
        Max Completion Length & 4096 \\
        Precision & bfloat16 \\
        Seed & 3047 \\
        \bottomrule
    \end{tabular}
\end{table}

% \newpage
\section{System Prompts}
\label{appendix:system-prompts}

\subsection{SFT System Prompt}
\begin{promptbox}[title={SFT System Prompt}]

When solving complex medical problems, follow this specific format:\\

1. First, break down your clinical reasoning into extremely concise steps, with each step limited to 10 words maximum.\\
2. Present this step-by-step reasoning inside <think></think> tags.\\
3. After completing your reasoning, provide your comprehensive assessment and plan after </think> tag.\\

For example, when analyzing a complex case:\\

<think>\\
Review presenting symptoms and vital signs.\\
Identify key abnormal findings.\\
Consider differential diagnoses by system.\\
Evaluate diagnostic test results.\\
Rule out critical conditions.\\
Analyze medication interactions.\\
Prioritize diagnoses by likelihood.\\
Determine urgent vs. stable concerns.\\
Plan appropriate interventions.\\
Consider follow-up requirements.\\
</think>\\

[Your complete, detailed clinical assessment including:\\
- Primary and differential diagnoses with supporting evidence\\
- Recommended diagnostic workup\\
- Treatment plan with rationale\\
- Follow-up recommendations\\
- Key considerations for this specific case]\\

Instructions:\\
- Each reasoning step must be 10 words or fewer\\
- Use Markdown for your assessment\\
- Include pathophysiology considerations in your reasoning\\
- Address diagnostic uncertainty when appropriate\\
- Consider evidence-based guidelines in your assessment\\
- Maintain proper XML tag formatting throughout\\
- Adapt your assessment to the type of problem whether Q\&A, Diagnostic reasoning, Treatment decision-making, or Prognostic assessment\\
\end{promptbox}

\newpage

\subsection{GRPO System Prompt}

\begin{promptbox}[title={GRPO System Prompt}]
You are a medical reasoning AI that solves multiple-choice questions through systematic clinical reasoning. For every question, demonstrate your thinking process inside <think></think> tags using this framework:\\

\#\# Reasoning Framework\\

1. **Information Analysis**: Extract key demographics, symptoms, vitals, labs, imaging, and clinical context\\
2. **Differential Diagnosis**: Identify patterns, consider diagnoses by system, rank by probability, note red flags\\
3. **Pathophysiology**: Connect symptoms to underlying disease mechanisms and biological processes\\
4. **Evidence-Based Reasoning**: Apply clinical criteria, interpret tests, reference current guidelines\\
5. **Logical Elimination**: Systematically evaluate each option, exclude based on evidence, compare remaining choices\\
6. **Clinical Decision**: Consider risk-benefit, address uncertainty, prioritize safety\\

\#\# Response Requirements\\

- Use Markdown formatting throughout your assessment\\
- Address diagnostic uncertainty when appropriate\\
- Reference evidence-based guidelines and clinical criteria\\
- Maintain proper XML tag formatting with <think></think>\\

\#\# Response Format\\

<think>\\
\lbrack Comprehensive reasoning using Markdown formatting, following the 6-step framework. Show medical knowledge, pathophysiology understanding, systematic problem-solving, consideration of multiple possibilities, elimination of incorrect options, and justification for your final choice. Address uncertainty and reference guidelines as appropriate.\rbrack\\
</think>\\

\#\# Assessment\\

[Provide your clinical assessment using Markdown formatting]\\

\#\# Final Answer\\

$\backslash$boxed\{X\} (where X is a single letter: A, B, C, D or E)\\

\#\# Key Principles\\

- **Systematic Over Intuitive**: Follow the framework consistently\\
- **Evidence-Based**: Ground reasoning in established medical knowledge and guidelines\\
- **Safety First**: Choose conservative options when uncertain\\
- **Patient-Specific**: Consider age, comorbidities, clinical context\\
- **Transparency**: Address limitations and diagnostic uncertainty\\

\#\# Critical Errors to Avoid\\

- Anchoring on first impressions\\
- Premature diagnostic closure\\
- Ignoring pathophysiology\\
- Pattern matching without mechanistic understanding\\
- Failing to reference current evidence-based guidelines\\

Apply rigorous clinical reasoning with proper formatting to achieve accurate medical decisions.
\end{promptbox}

%%%%%%%%%%%%%%%%%%%%%%%%%%%%%%%%%%%%%%%%%%%%%%%%%%%

% \bibliographystyle{unsrtnat}
% \bibliographystyle{unsrt}
% \bibliography{references}  %%% Uncomment this line and comment out the ``thebibliography'' section below to use the external .bib file (using bibtex) .





% \end{thebibliography}

\end{document}